\documentclass[review,authoryear]{elsarticle}

\usepackage{array}
\newcolumntype{P}[1]{>{\centering\arraybackslash}p{#1}}
\usepackage{lineno,hyperref}
\usepackage{graphicx}  % remove "demo" option in your real document
\usepackage[labelsep=quad,indention=10pt]{subfig}
\usepackage{caption}
\usepackage{ dsfont }
\usepackage{float}
\usepackage{soul}
\usepackage{multirow}
\usepackage{tabularx}
\usepackage[dvipsnames]{xcolor}
\usepackage[utf8]{inputenc}
\usepackage[noadjust]{cite}
\usepackage{amsmath}
\usepackage[capitalise]{cleveref}
\usepackage{float}
\usepackage[normalem]{ulem}
\usepackage{todonotes}

\makeatletter
\def\fixedlabel#1#2{%
  \@bsphack%
  \protected@write\@auxout{}%
         {\string\newlabel{#1}{{#2}{\thepage}}}%
  \@esphack}
\makeatother

\modulolinenumbers[5]

\usepackage{todonotes}

\makeatletter
\def\fixedlabel#1#2{%
  \@bsphack%
  \protected@write\@auxout{}%
         {\string\newlabel{#1}{{#2}{\thepage}}}%
  \@esphack}
\makeatother

\journal{Smart Agricultural Technology}

\begin{document}
\nolinenumbers

\begin{frontmatter}

\title{HOB-CNN: Hallucination of Occluded Branches with a Convolutional Neural Network for 2D Fruit Trees}

\author{Zijue Chen}
\author{Keenan Granland}
\author{Rhys Newbury}
\author{Chao Chen\corref{mycorrespondingauthor}}

\cortext[mycorrespondingauthor]{Corresponding author}
\ead{chao.chen@monash.edu}

\address{Laboratory of Motion Generation and Analysis, Faculty of Engineering, Monash University, Clayton, VIC 3800,Australia}

% \begin{document}

% \maketitle

\begin{abstract}

Orchard automation has attracted the attention of researchers recently due to the shortage of global labor force. To automate tasks in orchards such as pruning, thinning, and harvesting, a detailed understanding of the tree structure is required. However, occlusions from foliage and fruits can make it challenging to predict the position of occluded trunks and branches. This work proposes a regression-based deep learning model, Hallucination of Occluded Branch Convolutional Neural Network (HOB-CNN), for tree branch position prediction in varying occluded conditions. We formulate tree branch position prediction as a regression problem towards the horizontal locations of the branch along the vertical direction or vice versa. We present comparative experiments on Y-shaped trees with two state-of-the-art baselines, representing common approaches to the problem. Experiments show that HOB-CNN outperform the baselines at predicting branch position and shows robustness against varying levels of occlusion. We further validated HOB-CNN against two different types of 2D trees, and HOB-CNN shows generalization across different trees and robustness under different occluded conditions.

\end{abstract}
\end{frontmatter}

% \linenumbers

\section{Introduction}

Due to the global labor force shortage and the fast development of robots and artificial intelligence in recent years, automation in the horticulture industry has attracted attention from researchers. However, agricultural environments can be complex and unpredictable compared with industrial settings, making current agricultural tasks heavily dependent on manual labor. 

\begin{figure}[h!]
    \centering
    \includegraphics[width=11cm]{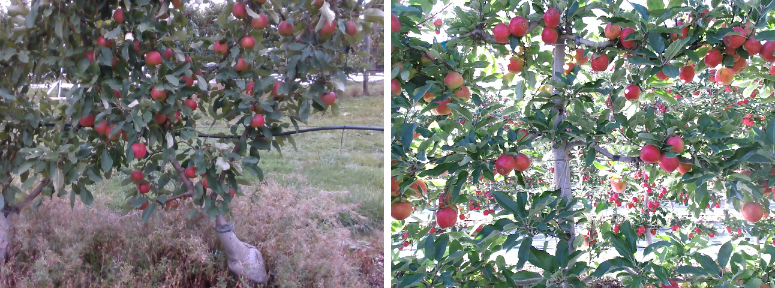}
    \caption{2D fruit wall structured trees. Y-shaped tree (left) and trellis tree (right).}
    \label{fig:intro2D}
\end{figure}

The automation of agricultural tasks requires the development of robotics, artificial intelligence, and agriculture. While more robust robots and vision systems are emerging, modern farms and orchards are also training the plants to be more automation friendly. For example, the 2D fruit wall growing patterns (e.g., trellis trees\footnote{Washington State University\\ https://research.libraries.wsu.edu/xmlui/handle/2376/17719} and Y-shaped trees, as shown in \cref{fig:intro2D}) has become a trend in commercial orchards in recent years. These planar tree structures simplify the automation of orchards and can benefit orchard maintenance. Therefore, current orchard automation research mainly focuses on 2D fruit wall structured trees.

Fruit orchard tasks, such as thinning, pruning, and harvesting, require an accurate tree skeleton model. However, fruit trees are often occluded by artifacts in the environment, such as foliage and fruits, which makes it a challenge for current computer vision technology~\citep{CHEN2021105952}. Furthermore, branch detection currently lacks a publicly available annotated datasets, unlike the more studied tasks of fruit detection, weed detection, and disease detection.

Branch detection is commonly done using semantic segmentation~\citep{9051680, ZHANG2018386, MAJEED201875, CHEN2021105952, LIANG2020105192, granland2021minimizing}. \citet{MAJEED201875} used SegNet~\citep{SegNet} and \citet{LIANG2020105192} used U-Net~\citep{unet} to segment the visible region of both branches and trunks of a tree. \citet{KANG2020105302} created a lightweight network, DasNet, to segment both apples and visible branches simultaneously. However, they are prone to produce large gaps in the output mask due to environmental occlusions. Furthermore, segmenting both visible and occluded branches has been shown to be a hard task for current neural network architectures~\citep{CHEN2021105952}. \citet{granland2021minimizing} develop a semi-supervised approach, where an automated process would try to repair the label generated by U-Net. 

To recover information from the gaps in the output mask, post-processing steps are used to hallucinate the occluded branches based on the previous detection result. \citet{ZHANG2018386} and \citet{AMATYA20163} fit a polynomial curve to the semantic segmentation result, aiming to fill the gaps in the tree skeleton. However, this is prone to errors in false positive branch detection results. \citet{amatya2016integration} detected the visible cherry stem and cherry clusters during the night, then used geometric relations to infer the occluded stem. \citet{MAJEED2020105308,MAJEED2020105671} trained a ResNet~\citep{resnet34} based object detection neural networks to detect visible branches with bounding boxes. Curve fitting was applied to the center of these bounding boxes for occluded grapevine detection. This was extended in \citep{Majeed2021} where they applied semantic segmentation to leaves that occluded the grapevines. A geometric relationship was then used to deduce the occluded section of the grapevine. However, these hallucination-based methods are completely dependent on the detection results from previous steps and post processing steps commonly do not make use of available information from the original images.

Previous work complete branch hallucination in at least two steps, in comparison, humans have achieve this autonomously, understanding that an object exists even if occluded. This is known as `object permanence' and research has shown humans as young as 3-month-old~\citep{bower1974development,baillargeon1991object} exhibit this trait. Based on the visible branches and the prior knowledge of the spatial structure, humans can deduce the coherent tree branches regardless of the occlusions. We aim to translate this intuition into a deep learning model, training it to predict both visible and occluded branches in a single step. 

This work proposes a novel regression-based deep learning model, Hallucination of Occluded Branch Convolutional Neural Network (HOB-CNN), to predict the branch position of 2D fruit trees under different occlusion conditions in one step. HOB-CNN guarantees completely connected branch prediction helping to remove noise and prevent gaps in the predictions. The accuracy of HOB-CNN is compared to two current state-of-the-art (SOTA) methods. The experiments illustrated that HOB-CNN has higher accuracy in branch position prediction, and good robustness against occlusions. HOB-CNN is tested on three datasets and it shows good generalization and robustness across different tree structures and occlusion conditions.

To our best knowledge, this is the first application of an end-to-end regression-based deep learning model to predict tree position considering occlusions. The contributions of this work are threefold: 
\begin{itemize}
    \item A novel one-step regression-based deep learning model, HOB-CNN, to predict tree branch position of 2D fruit trees under occluded conditions.
    \item Comparative experiments between HOB-CNN and two SOTA methods in tree position prediction.
            \item An annotated tree position prediction dataset of Y-shaped trees in three different seasons. The dataset can be accessed \href{https://drive.google.com/file/d/1-6akVXezq26ihtDrVH9czGUQDFd8Py81/view?usp=sharing}{here}.
\end{itemize}

The organization of this work is as follows: the utilised datasets are introduced in \cref{sec:datasets}, the detailed methodology is described in \cref{sec:Methodology}, the results and discussion are shown in \cref{sec:r+d} before concluding in \cref{sec:conclusion}.

\section{Datasets}
\label{sec:datasets}

Three datasets were used in this work (\cref{table:threedatasets}). Firstly, a dataset containing RGB-D images of Y-shaped trees acts as the main dataset used for experiments in this work. This will be referred to as `Y-shaped Tree Dataset', which includes images from different seasons, where each season will have a varying amount of environmental occlusion. We also test our approach on a dataset of trellis trees (referred to as `Trellis Tree Dataset') and a dataset of grapevines (referred to as `Grapevine Dataset'). Both of these datasets consist of RGB images, and are used to test the generalization of HOB-CNN. The original unlabeled images of these two datasets are from AgRobotics Laboratory\footnote{Washington State University\\ https://research.libraries.wsu.edu/xmlui/handle/2376/17719}. Y-shaped tree and trellis tree are two prevalent 2D tree structures used in modern orchards, and grapevines are typical horizontally grown plants. These three datasets aim to provide diversity in type, structure and occlusion conditions.

\begin{table}[h!]
\centering
\begin{tabular}{cccc}
\hline
\textbf{Tree Type} & \textbf{Image Number} & \textbf{Image Type} & \textbf{Source}       \\ \hline
Y-shaped Tree      & 2178                      & RGB-D               & Self                \\
Trellis Tree       & 575                       & RGB                 & AgRobotics Laboratory \\
Grapevine          & 434                       & RGB                 & AgRobotics Laboratory \\ \hline
\end{tabular}
\caption{Utilized tree datasets in this work.}
\label{table:threedatasets}
\end{table}

All the images were center cropped to square and resized to 256 $\times$ 256 pixels for neural networks training purpose.

\subsection{Y-shaped Tree}

\begin{figure}[h!]
    \centering
    \includegraphics[width=11cm]{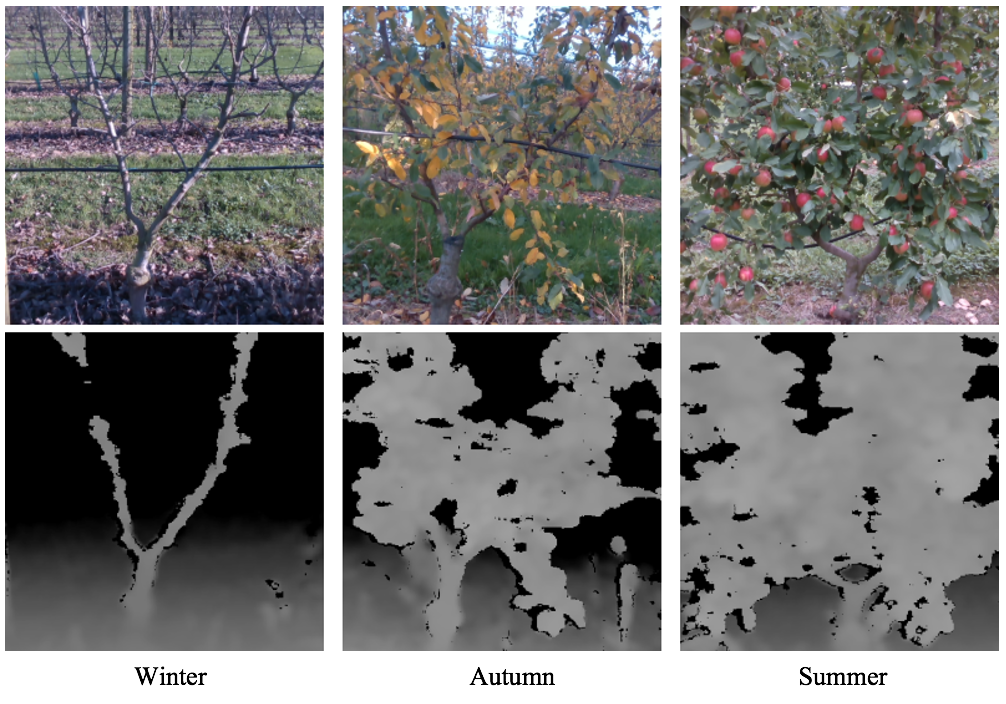}
    \caption{Example of Y-shaped trees in different seasons. RGB images on top, depth maps on bottom.}
    \label{fig:seasonRGBD}
\end{figure}

Y-shaped Tree Dataset includes $1099$ winter, $518$ autumn and $561$ summer RGB-D images from a commercial apple orchard in northeastern Melbourne, VIC. The apple species is Ruby Pink and the trees are planted east to west, with 4.125 m between the centers of each row of trees and 0.6 m between each tree in the row. A Intel RealSense D435 was used to capture RGB-D images, and the distance between the camera and the trees is between 1.8 m and 2.2 m. The trees in winter images have no occlusions from the external environment. Autumn images were collected after harvest season, when the leaves begin to fall off. Summer images were taken during harvest season, where occlusions are from both foliage and fruits. The occluded branch percentage distribution of the trees in three seasons are shown in \cref{table:seasonocclusion}, and an example from each season are shown in \cref{fig:seasonRGBD}. We aim at predicting the position of trunk and primary branches (the branches that directly attach to the trunk) for Y-shaped trees.

\begin{table}[h!]
\centering
\begin{tabular}{cccc}
\hline
\textbf{Season} & \textbf{Image Number} & \textbf{Occlusion Percentage} & \textbf{Occlusion Amount} \\ \hline
Winter          & 1099                  & $0\pm0$                           & None                   \\
Autumn          & 518                   & $14\pm15$                         & Medium                    \\
Summer          & 546                   & $36\pm9$                          & Heavy            \\ \hline
\end{tabular}
\caption{Occlusion details of Y-shaped Tree Dataset in winter, autumn and summer.}
\label{table:seasonocclusion}
\end{table}

The images from all seasons were randomly divided into five cross-validation dataset groups evenly, and \cref{table:Ytree_divid} shows the details of each group. In each cross validation step, four groups of images were used for training and one group was used for testing. For instance, in Trial 1 the test dataset consists of Group 1 and the training dataset consists of Group 2-5.

\begin{table}[h!]
\centering
\begin{tabular}{cccc}
\hline
\textbf{Group Number} & \textbf{Summer} & \textbf{Autumn} & \textbf{Winter} \\ \hline
1                     & 103             & 89              & 243             \\
2                     & 132             & 103             & 201             \\
3                     & 115             & 95              & 225             \\
4                     & 113             & 109             & 214             \\
5                     & 98              & 122             & 216             \\ \hline
Total                 & 561             & 518             & 1099            \\ \hline
\end{tabular}
\caption{The compositions of cross validation groups in Y-shaped Tree Dataset.}
\label{table:Ytree_divid}
\end{table}

\begin{figure}[h!]
    \centering
    \includegraphics[width=\textwidth]{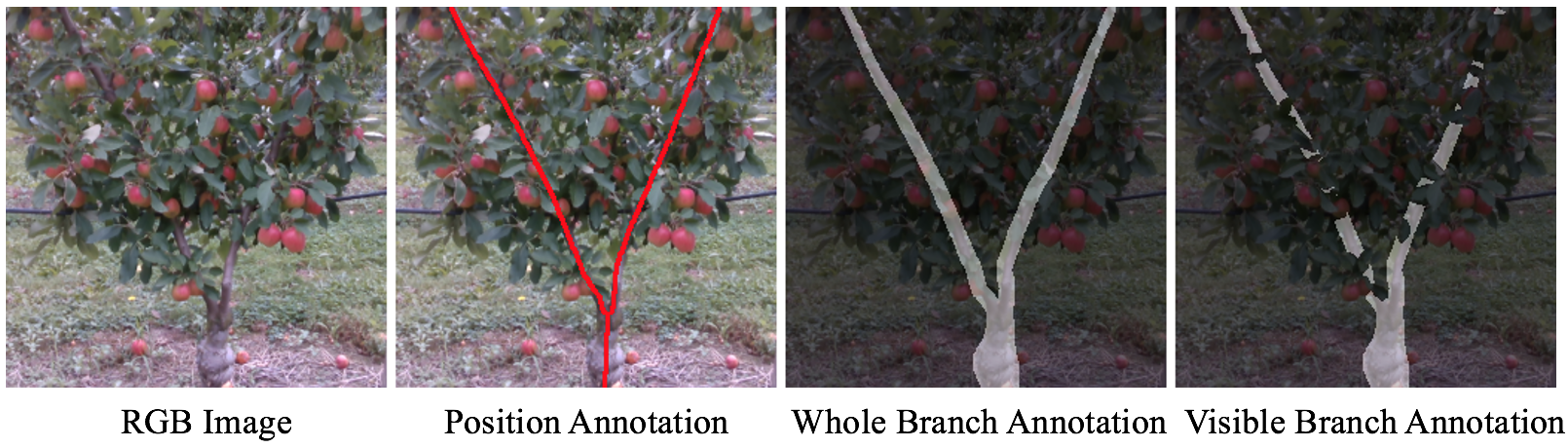}
    \caption{Example of annotations for Y-shaped Tree Dataset}
    \label{fig:Yannotation}
\end{figure}

The annotation process for Y-shaped Tree Dataset are as follows
(\cref{fig:Yannotation}):

\begin{enumerate}
  \item The \textit{Position Annotation} (red line) is manually labeled on the RGB images.
  \item Scanning from top to bottom across the entire \textit{Position Annotation}, for each vertical pixel row, the horizontal center locations of both the left and right branch are recorded as the position ground truth. After the branches join together in trunk segment, the only location is assigned to both left and right branches. The final HOB-CNN training target of each Y-shaped tree image is an array with the shape of $2h$, where $h$ is the height of the image.
  \item The pixel-by-pixel annotation of the trunk and primary branches was drawn on the RGB image, with the \textit{Position Annotation} serves as reference, which is called \textit{Whole Branch Annotation}. 
  \item The pixels that belonging to occlusions were removed from \textit{Whole Branch Annotation}, and what left is \textit{Visible Branch Annotation}. 
\end{enumerate}

\subsection{Trellis Tree}

Trellis tree (also called espalier tree or cordon tree) are when trees are manipulated to grow along a trellis, forming a 2D fruit wall structure. The original images of the Trellis Tree Dataset were from ‘Scifresh Apple Original and DepthFilter RGB Images’\footnote{http://hdl.handle.net/2376/17720}. This dataset includes 575 images of trellis apple trees in different levels of blurriness and lighting conditions. The 575 trellis tree images were divided into five groups for cross validation, with 115 images in each group.

\begin{figure}[htp]
    \centering
    \includegraphics[width=\textwidth]{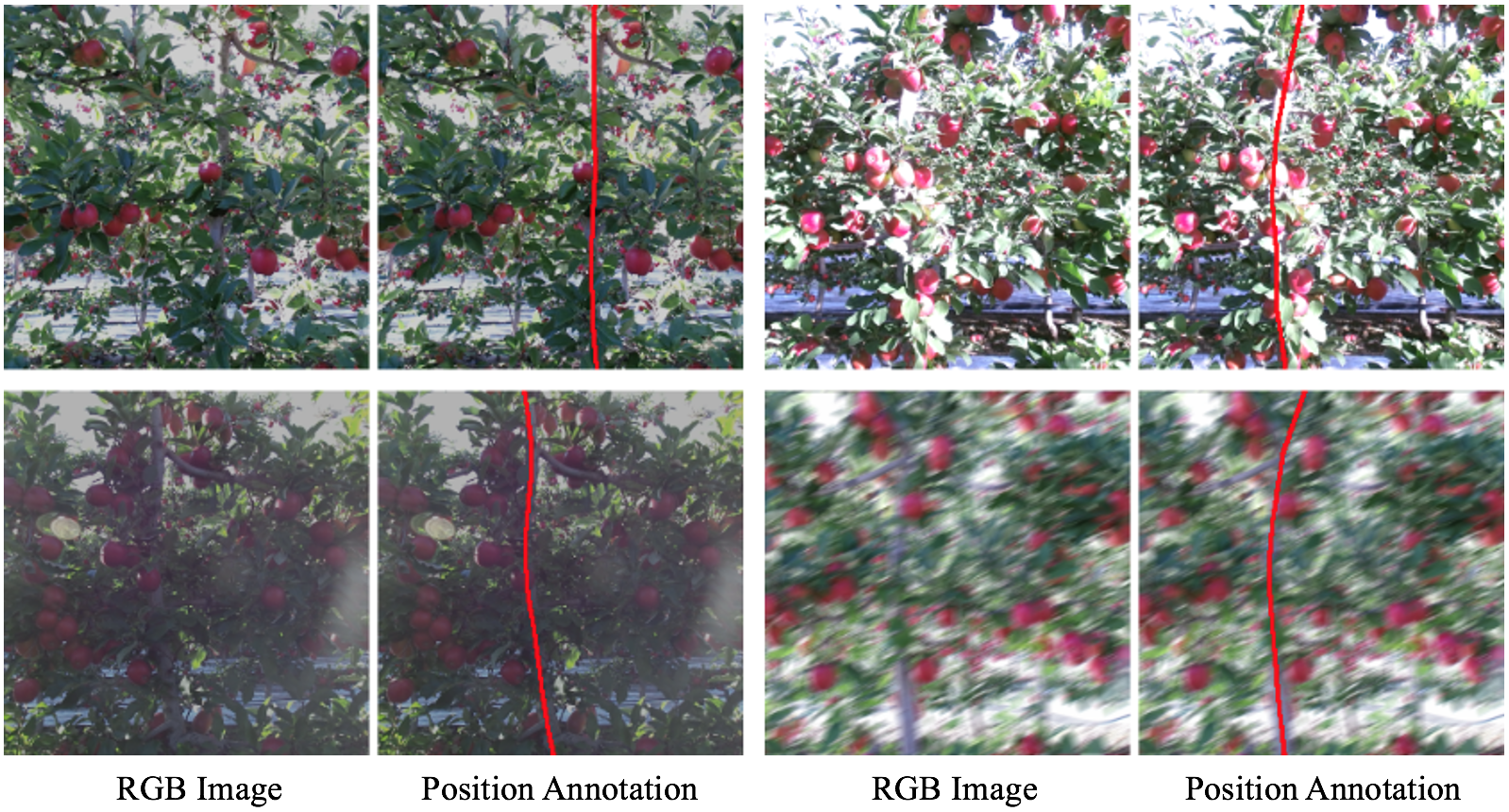}
    \caption{Annotation examples for Trellis Tree Dataset}
    \label{fig:Tannotation}
\end{figure}

For Trellis Tree Dataset (\cref{fig:Tannotation}), the horizontal branches of the trellis trees are heavily occluded by foliage and fruits, which makes it very difficult for humans to annotate accurately. Thus, this work focuses on only tree trunk prediction. The generation of HOB-CNN training targets follows the same process as Y-shaped Tree corresponding to the \textit{Position Annotation}.

\subsection{Grapevine}

\begin{figure}[h!]
    \centering
    \includegraphics[width=\textwidth]{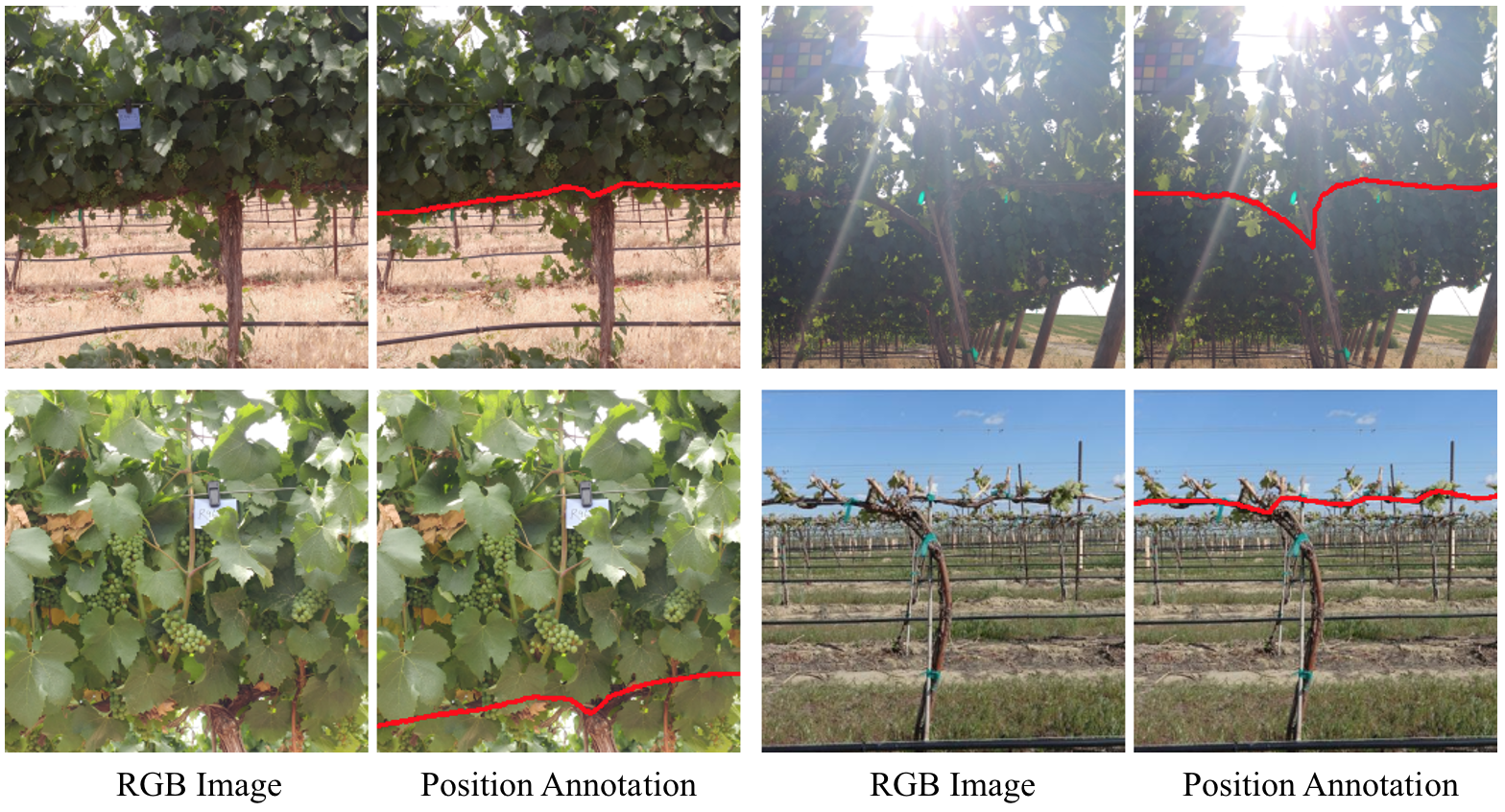}
    \caption{Annotation examples for Grapevine Dataset}
    \label{fig:Gannotation}
\end{figure}

Grapevine Dataset includes images of grapevines from early spring and summer (\cref{fig:Gannotation}). The original 184 grapevine images in early spring are from `Vineyard Canopy Images during Early Growth' Stage\footnote{http://hdl.handle.net/2376/16939}, and 250 original summer grapevine images are from `Full Stages of Wine Grape Canopy and Clusters'\footnote{http://hdl.handle.net/2376/17628}. The grapevine images from early spring have minimal occlusions from the early stage shoots. Summer grapevines are heavily occluded by leaves. Additionally, the summer dataset contains images with different camera angles, lighting conditions and maturity, which increased the diversity of summer images.

To make up the limited training images of the Grapevine Dataset and generate more training images, an augmentation process was applied to each image, square cropping from top, center and bottom with a side length equal to $0.8\times H$, where $H$ is the height of the original images. 

Then, all the cropped images were distributed into five cross validation groups, the details of each group are shown in \cref{table:G_divid}.

\begin{table}[h!]
\centering
\begin{tabular}{ccc}
\hline
\textbf{Group Number} & \textbf{Early Sprint} & \textbf{Summer} \\ \hline
1                     & 114                   & 147             \\
2                     & 99                    & 162             \\
3                     & 114                   & 147             \\
4                     & 113                   & 147             \\
5                     & 112                   & 149             \\ \hline
Total                 & 552                   & 752             \\ \hline
\end{tabular}
\caption{The compositions of cross validation groups in Grapevine Dataset}
\label{table:G_divid}
\end{table}

\section{Methodology}
\label{sec:Methodology}

\subsection{HOB-CNN}
\label{sec:hobcnn}

HOB-CNN was designed to predict the horizontal location of the branch along a vertical direction, or vice versa. This is applicable to both vertical trees (e.g. Y-shaped trees, trellis trees) and horizontal vines (e.g. grapevines). However, if the tree has multiple branches, there will be multiple locations. Therefore, HOB-CNN will output one horizontal location for each branch at each pixel row. For example, the prediction of Y-shaped trees has two horizontal locations in each vertical pixel row, one for the left branch and one for the right branch. 

\begin{figure}[h!]
\centering
\includegraphics[width=\textwidth]{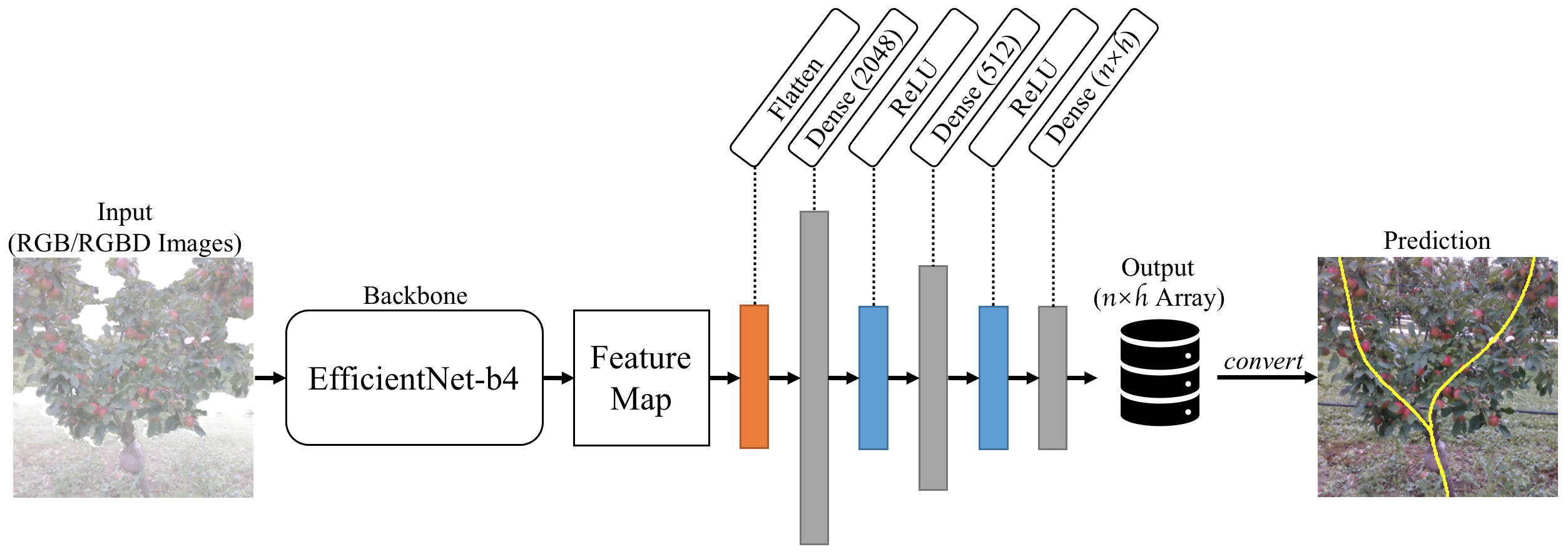}
\caption{HOB-CNN architecture}
\label{fig:HOBCNN architecture}
\end{figure}

HOB-CNN consists of a backbone, followed by three fully connected layers. We used EfficientNet-b4~\citep{tan2020efficientnet} as the backbone, which has SOTA performance on image recognition with a smaller amount of parameters compared to similarly performing networks. The output of the backbone was flattened, then fed into dense layers (see \cref{fig:HOBCNN architecture}). The first and second dense layers have 2048 and 512 neurons, respectively, with ReLU activation. Finally, the last dense layer outputs $n\times h$ neurons, where $n$ is the number of branches in the tree, and $h$ is the height/width of the image in pixels.

\subsection{Baselines: U-Net with Curve Fitting}

\textbf{U-Net}~\citep{unet} is a fully convolutional neural network for semantic segmentation. In our previous work~\citep{CHEN2021105952}, it was found that U-Net outperformed other SOTA methods at predicting occluded tree branches. Therefore we used the same U-Net model, with ResNet-34~\citep{resnet34} as the encoder. U-Net was used to train two different models.

\begin{enumerate}
    \item \textbf{U-Net Visible} to detect only \textit{visible} branch sections
    \item \textbf{U-Net Whole} to hallucinate the \textit{whole} branch under occlusions.
\end{enumerate}

\textbf{Curve Fitting} is a common post-processing method in branch detection (e.g \citep{ZHANG2018386,MAJEED2020105308,MAJEED2020105671,Majeed2021}), and was utilized in this work to create a continuous curve from a possibly noisy semantic mask from \textit{U-Net Visible} or \textit{U-Net Whole}. In this work, `\textit{U-Net Visible} + curve fitting' method is referred to as \textit{Visible CF}, and `\textit{U-Net Whole} + curve fitting' is referred to as \textit{Whole CF}.

\begin{figure*}[htp]
    \centering
    \includegraphics[width=\textwidth]{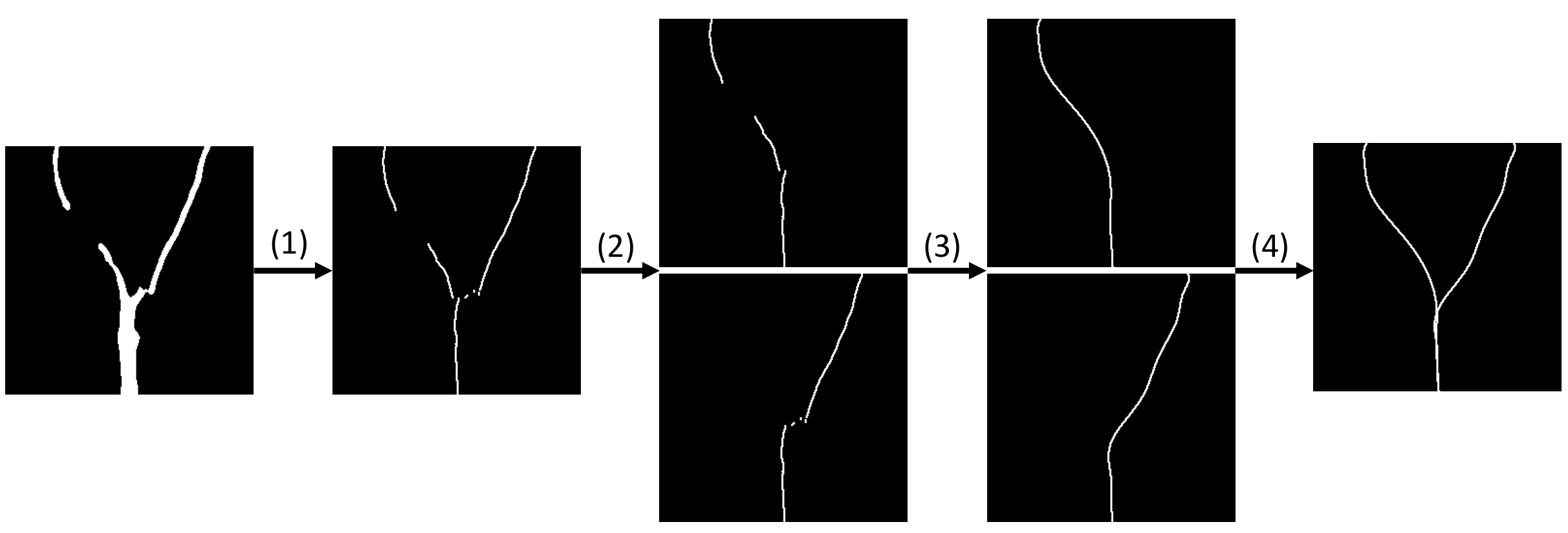}
    \caption{U-Net curve fitting procedure: (1) remove noise and extract waypoints; (2) split waypoints to 'Left' and 'Right'; (3) fit curve on waypoints; (4) combine fitted curves together.}
    \label{fig:Unetcurvefit}
\end{figure*}

The semantic predictions from the two U-Net models were converted to a single-pixel width skeleton in a multi-step process. Firstly, filters were used to reduce noise, followed by fitting a curve to the result. Two curve fitting methods were tested: cubic splines and $n$th polynomial (3rd to 15th). The 5th order polynomial curve fitting was used as it had the best performance. The process is as follows (\cref{fig:Unetcurvefit}):

\begin{itemize}
  \item A blob filter was applied removing blobs with an area less than 65 pixels.
  \item Waypoints were calculated for curve fitting by converting the binary label to a series of points, taking the center blob position of each row.
  \item The waypoints were then split into a 'Left' and 'Right' path. 
  \item After the 'Left' and 'Right' path were generated, a 5th order polynomial curve was fitted to each path and used to generate the output label.
\end{itemize}

\subsection{Training Details}

\textit{U-Net Visible} and \textit{U-Net Whole} were trained on the three datasets separately (RGBD input images for Y-shaped Tree Dataset and RGB input images for Terllis Tree and Grapevine Datasets), with a batch size of 8 for 50 epochs, and a horizontal flip was used in augmentation. Weighted dice loss~\citep{wdl} and the Adam optimizer were utilized. 

Similarly, HOB-CNN was trained by each dataset separately with a batch size of 8, and horizontal flip was used in augmentation. Mean Squared Error (L2 loss) was used as the loss function and the Adam optimizer was utilized. The model was evaluated after convergence and before overfitting was observed, which is 90 epochs for Y-shaped Trees Dataset, 400 epochs for Trellis Trees Dataset and 300 epochs for Grapevine Dataset.

\subsection{Metrics}
\label{sec:metrics}

 Root Mean Squared Error (RMSE) and Correlation Coefficient (r) were adopted for evaluating the accuracy of the predicted branch positions of \textit{Visible CF}, \textit{Whole CF} and HOB-CNN.
 
\textbf{Root Mean Squared Error (RMSE)}: RMSE was used in this work to measure the error between the tree position prediction and ground truth:

\begin{equation}
    RMSE = \sqrt{\frac{\sum_{n=1}^{N} (g_i - p_i)^2}{N}}
\end{equation}
where $N$ is the number of data points, $g_i$ and ${p_i}$ are the ground truth value and the predicted value respectively.

\textbf{Correlation Coefficient (r)}: Correlation coefficient is a widely used metric for regression analysis. It is defined as the degree of relationship between two groups of data. Correlation coefficient is evaluated between $[-1,1]$, where $1$ means the two variables have perfect correlation and $-1$ means the two variables are in perfect opposites, $0$ indicates they are not related. The equation is as follows:
\begin{equation}
    r = \frac{\sum(p_i-\Bar{p})(g_i-\Bar{g})}{\sqrt{\sum(p_i-\Bar{p})^2\sum(g_i-\Bar{g})^2}}
\end{equation}
where $p_i$ and $\Bar{p}$ are the predicted values and the mean of the predicted values; $g_i$ and $\Bar{g}$ are the ground truth values and the mean of the ground truth values.

\section{Results and Discussions}
\label{sec:r+d}

In \cref{sec:hobcnn_baselines}, we first discussed the performance of HOB-CNN at detecting the branch positions of Y-shaped trees and compared its results with \textit{Visible CF} and \textit{Whole CF}. We analyzed the impact of occlusion on their performance in \cref{sec:performance v occlusion}. In \cref{sec:time_comparison}, processing time and annotation time of each method are compared. HOB-CNN is also applied on branch detection of trellis trees and grapevines in \cref{sec:other datasets}. Finally, we analyzed the common features of HOB-CNN predictions with large errors in \cref{sec:error}.

\subsection{Y-shaped Tree Position Prediction}
\label{sec:hobcnn_baselines}

\begin{figure}[h!]
    \centering
    \includegraphics[width=\textwidth]{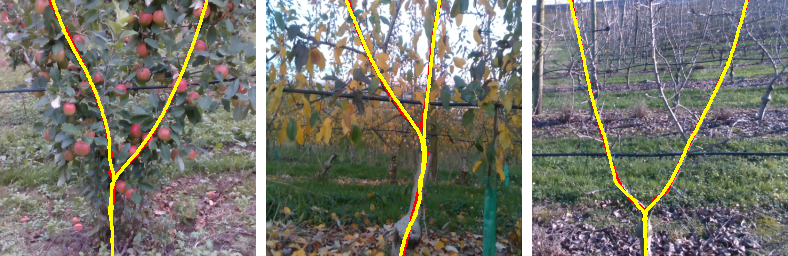}%, trim=0 0 5.7cm 0, clip
    \caption{HOB-CNN prediction examples on Y-shaped tree at summer (left), autumn (middle) and winter (right). Red lines represent the GT.}% 
    \label{fig:firstdemo}
\end{figure}

In the experiment of Y-shaped tree prediction, HOB-CNN shows good performance at hallucinating complete branch paths under different occlusions. Some examples are shown in \cref{fig:firstdemo}, where the yellow curves represent branch center lines that HOB-CNN predicted while the underling red curves are the ground truth.

To quantify the accuracy of HOB-CNN, RMSE and correlation coefficient are utilized for evaluation. The results in different seasons are compared with the two baselines,  \textit{Visible CF} and \textit{Whole CF}. As shown in \cref{table:threecompare}, HOB-CNN has the highest accuracy in all seasons. This advantage is more evident in summer when the occlusion is heavy: the mean RMSE of HOB-CNN in summer is 2.671 pixels, obviously lower than the widely used \textit{Visible CF} (5.651 pixels) and \textit{Whole CF} (4.151 pixels). Comparing \textit{Visible CF} and HOB-CNN, the standard deviations of RMSE and correlation coefficient drops from 6.235 to 1.589 and from 0.092 to 0.032 respectively. This shows that HOB-CNN’s performance is more stable under different occlusion scenarios.

\begin{table}[h!]
\centering
\begin{tabular}{ccccc}
\hline
\textbf{Metric}                                                                              & \textbf{Season} & \textbf{Visible CF}                 & \textbf{Whole CF}                   & \textbf{HOB-CNN}                             \\ \hline
                                                                                             & Winter          &2.622±4.020 &2.719±3.741 &\textbf{1.962±1.290} \\
                                                                                             & Autumn          &3.312±4.546 &3.421±6.839 &\textbf{2.463±2.023} \\
                                                                                             & Summer          &5.651±9.705 &4.151±5.755 &\textbf{2.671±1.549} \\
\multirow{-4}{*}{\textbf{RMSE}}                                                              & Total           &3.566±6.235 &3.255±5.201 &\textbf{2.264±1.589} \\ \hline
                                                                                             & Winter          &0.992±0.055 &0.992±0.068 &\textbf{0.996±0.020} \\
                                                                                             & Autumn          &0.979±0.093 &0.979±0.097 &\textbf{0.989±0.053} \\
                                                                                             & Summer          &0.969±0.138 &0.984±0.073 &\textbf{0.995±0.024} \\
\multirow{-4}{*}{\textbf{\begin{tabular}[c]{@{}c@{}}Correlation\\ Coefficient\end{tabular}}} & Total           &0.983±0.092 &0.987±0.077 &\textbf{0.994±0.032} \\ \hline
\end{tabular}
\caption{RMSE and correlation coefficient of \textit{Visible CF}, \textit{Whole CF} and HOB-CNN in Y-shaped tree position prediction.}
\label{table:threecompare}
\end{table}

\begin{figure}[h!]
    \centering
    \includegraphics[width=\textwidth]{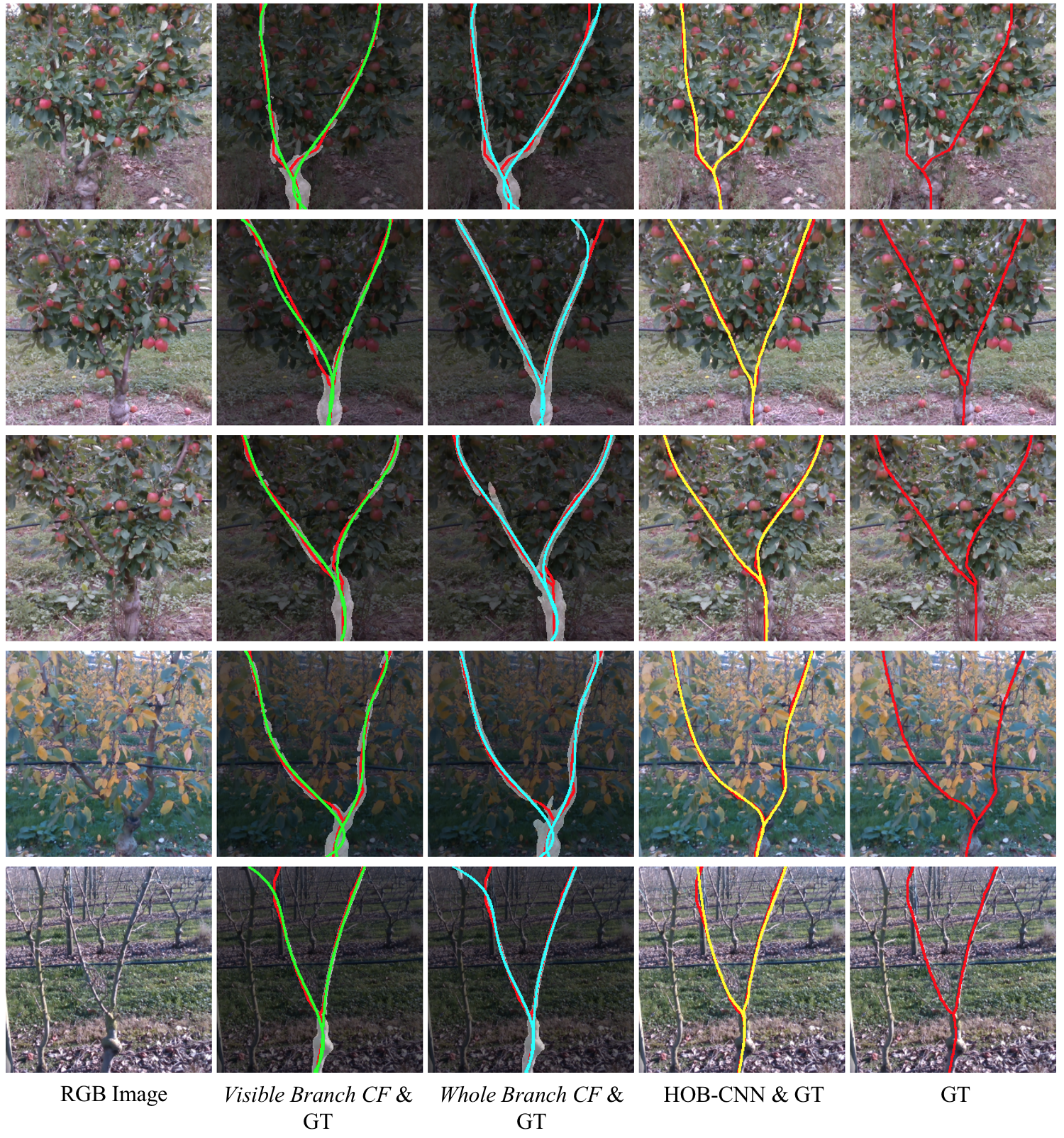}%, trim=0 0 5.7cm 0, clip
    \caption{Examples of \textit{Visible CF}, \textit{Whole CF} and HOB-CNN at Y-shaped tree position prediction. The red curves represent the ground truth annotation.}% 
    \label{fig:threecompare}
\end{figure}

Figure \ref{fig:threecompare} shows some examples of \textit{Visible CF}, \textit{Whole CF} and HOB-CNN at tree position prediction. The semantic predictions of visible branch and whole branch are also shown in the images. The semantic predictions of \textit{U-Net Whole} shows its potential of hallucinating occluded branches across various occlusion conditions, but can not guarantee the fully connectivity of the tree structure. However, \textit{U-Net Whole} tries to deduce the occluded branch segments with the visible branches as reference, and the accurate hallucinations offer more reference information for curve fitting, which can potentially explain the improved accuracy from \textit{Visible CF} to \textit{Whole CF} during summer in \cref{table:threecompare}.

It can also be seen from \cref{fig:threecompare} that the semantic predictions of both \textit{U-Net Visible} and \textit{U-Net Whole} in summer and autumn are more prone to noises and disconnections because of occlusions. These noises and disconnections introduce errors and lead to deviations in curve fitting results. On the other hand, 
low order curves cannot represent sharp bends accurately, whereas high order curves are prone to overfitting and are likely to generate too many bends. Even though in our experiments, a 5th order polynomial curve shows the best fitting results, the details of the predicted tree branches are also limited by the curves. For example, the sharp bend that happens at the merging point as shown in \cref{fig:threecompare} (top), cannot be accurately represented using a 5th order polynomial. In contrast, HOB-CNN infers branch positions directly on the original image, without using curve fitting. Therefore, HOB-CNN has more complete reference information, which increases the prediction accuracy. Also, as HOB-CNN is not restricted by curve fitting, it is more flexible at predicting branch structure, and can better fit the branch path where sharp bend occurs.

\subsection{Relationship between Performance and Branch Occlusion}
\label{sec:performance v occlusion}

\begin{figure}[h!]
    \centering
    \includegraphics[width=\textwidth]{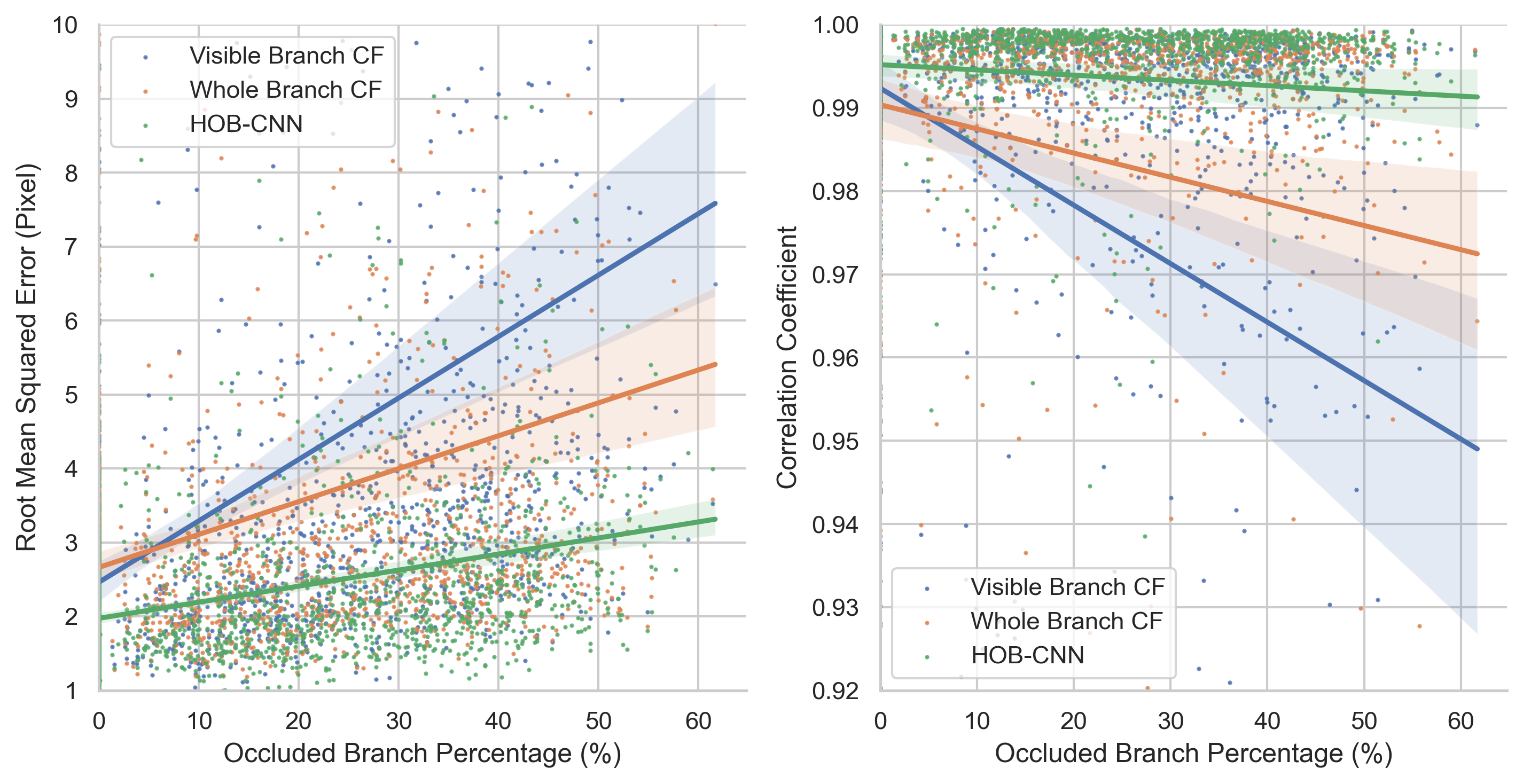}
    \caption{RMSE and correlation coefficient on varying Occluded Branch Percentage. The lines show the fitted trend of the data while the highlighted area represents the variance.}
    \label{fig:occlusion_rmse_r}
\end{figure}

To better demonstrate the influence of occlusion on tree position prediction, we show the trend of RMSE and correlation coefficient of \textit{Visible CF}, \textit{Whole CF} and HOB-CNN with increasing occlusion in \cref{fig:occlusion_rmse_r}. \textit{Visible CF} is the most vulnerable to occlusion. It outperforms \textit{Whole CF} under no occlusions, but the prediction accuracy decreases rapidly as the occlusion increases, and is surpassed by \textit{Whole CF} at 5\% occlusion. HOB-CNN has the most robust performance, consistently outperforming the two baselines in both RMSE and correlation coefficient with the smallest error and variance throughout.

\subsection{Processing and Annotation Time Comparison}
\label{sec:time_comparison}

\cref{table:operationtime} summarises the average processing time and annotation time for all three methods. \textit{Visible CF} and \textit{Whole CF} require post processing steps like filtering noises, extracting way points, as well as fitting curves. These extra steps add to the overall processing time. The neural network predicting times of both \textit{Visible CF} and \textit{Whole CF} are very close to that of HOB-CNN's, while most of the processing time is spent on curve fitting. The total processing times of \textit{Visible CF} and \textit{Whole CF} are 1451ms and 1716ms respectively. This is because \textit{Whole CF} needs to extract more way points for curve fitting than \textit{Visible CF}, so that the curve fitting time increases by 19\%. HOB-CNN, in comparison, only takes 58ms, about 3\% of the two baselines, giving it advantage for fast detection in field.

\begin{table}[h!]
\centering
\begin{tabular}{lccc}
\hline
\textbf{Model} & \textbf{Visible CF} & \textbf{Whole CF} & \textbf{HOB-CNN} \\ \hline
\multicolumn{4}{c}{\textbf{Processing Time (ms)}}                                            \\ \hline
Neural Network & 53                         & 52                       & 58               \\
Curve Fitting  & 1397                       & 1663                     & -                \\
Total          & 1451                       & 1716                     & 58               \\ \hline
\multicolumn{4}{c}{\textbf{Annotation Time (s)}}                                          \\ \hline
Winter         & 90                         & 90                       & 15               \\
Autumn         & 160                        & 120                      & 15               \\
Summer         & 210                        & 150                      & 20               \\
Average        & 153                        & 120                      & 17               \\ \hline
\end{tabular}
\caption{Processing and annotation times for \textit{Visible CF}, \textit{Whole CF} and HOB-CNN.}
\label{table:operationtime}
\end{table}

The required effort of ground truth labeling is another important consideration for model choosing. \textit{Visible CF} and \textit{Whole CF} require the semantic annotation, whereas HOB-CNN only requires the center line to represent the branch positions, so HOB-CNN saves labeling time greatly. Besides, since \textit{Visible CF} requires identifying non-occluded branch segments and accurately labeling them, it takes extra time than \textit{Whole CF}. From the table, The annotation time of HOB-CNN is 17s per image averagely, compared with \textit{Visible CF}'s 153s and \textit{Whole CF}'s 120s, it saves more than 85\% annotation time. Especially for summer trees, changing the model from \textit{Visible CF} to HOB-CNN makes the annotation time drop from 210s to 20s per image.

\subsection{HOB-CNN on Other Datasets}
\label{sec:other datasets}

We also apply HOB-CNN on predicting the trellis tree trunks and grapevine branches. By changing the number of neurons in the output layer, HOB-CNN can be applied to different types and structures of 2D trees. For example, the number of neurons in the output layer of the trellis trees model reduces neuron from 512 to 256 compared to the Y-shaped tree model. 

\begin{table}[h!]
\centering
\begin{tabular}{ccc}
\hline
\multirow{2}{*}{\textbf{Trial}} & \multicolumn{2}{c}{\textbf{Metrics}}        \\ \cline{2-3} 
                                & \textbf{RMSE}        & \textbf{Correlation Coefficient}           \\ \hline
1                               & 1.398±0.955 & 0.951±0.081          \\
2                               & 1.831±0.845          & 0.938±0.100          \\
3                               & 1.530±0.842          & 0.960±0.062 \\
4                               & 1.463±0.786          & 0.948±0.122          \\
5                               & 1.927±0.945          & 0.945±0.073          \\ \hline
Average                         & 1.630±0.902          & 0.948±0.091          \\ \hline      
\end{tabular}
\caption{Performance of HOB-CNN on trellis tree trunk detection}
\label{table:trellis}
\end{table}

\begin{figure}[h!]
    \centering
    \includegraphics[width=\textwidth]{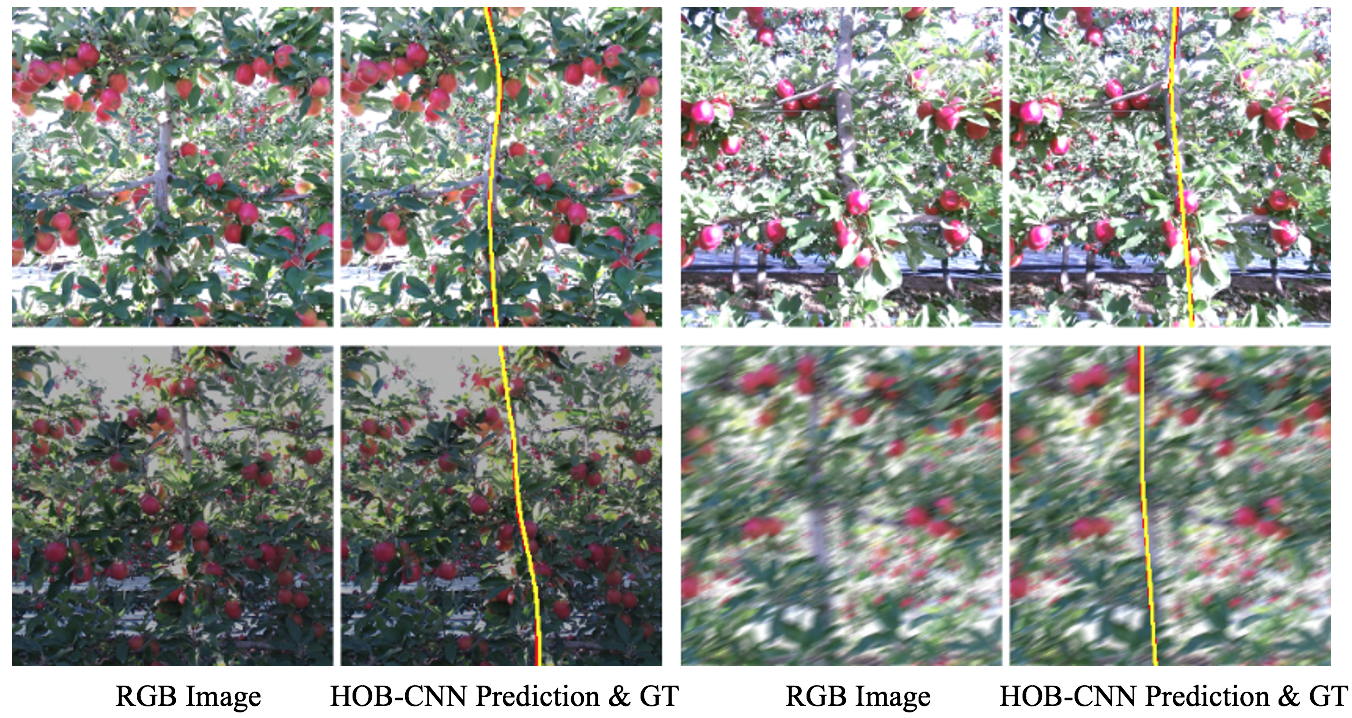}
    \caption{HOB-CNN prediction examples on Trellis Tree Dataset. The left images are RGB image and the right images are the prediction (yellow curves) overlay with ground truth (red curves).}
    \label{fig:trellis}
\end{figure}

\textbf{Trellis Tree Dataset}: ~\cref{table:trellis} illustrates that HOB-CNN has a consistent performance in detection of trellis tree trunks. Some prediction examples are shown in \cref{fig:trellis}. The results show the robustness of HOB-CNN under different lighting and blurriness conditions. 

\begin{table}[h!]
\centering
\begin{tabular}{ccccc}
\hline
\multirow{2}{*}{\textbf{Trial}} & \multicolumn{2}{c}{\textbf{Early Spring}}                                                  & \multicolumn{2}{c}{\textbf{Summer}}                                                        \\ \cline{2-5} 
                                & \textbf{RMSE} & \textbf{\begin{tabular}[c]{@{}c@{}}Correlation\\ Coefficient\end{tabular}} & \textbf{RMSE} & \textbf{\begin{tabular}[c]{@{}c@{}}Correlation\\ Coefficient\end{tabular}} \\ \hline
1                               & 1.759±0.606   & 0.946±0.035                                                                & 2.123±1.158   & 0.964±0.028                                                                \\
2                               & 2.794±1.077   & 0.862±0.140                                                                & 3.282±1.793   & 0.895±0.152                                                                \\
3                               & 1.869±0.880   & 0.950±0.040                                                                & 2.461±1.766   & 0.948±0.113                                                                \\
4                               & 2.706±3.892   & 0.921±0.057                                                                & 2.706±0.966   & 0.933±0.062                                                                \\
5                               & 2.337±1.439   & 0.935±0.052                                                                & 2.672±1.618   & 0.947±0.063                                                                \\ \hline
Average                         & 2.279±2.032   & 0.925±0.079                                                                & 2.661±1.554   & 0.937±0.099                                                                \\ \hline
\end{tabular}
\caption{Performance of HOB-CNN on grapevine branch detection}
\label{table:grapevine}
\end{table}

\begin{figure}[h!]
    \centering
    \includegraphics[width=\textwidth]{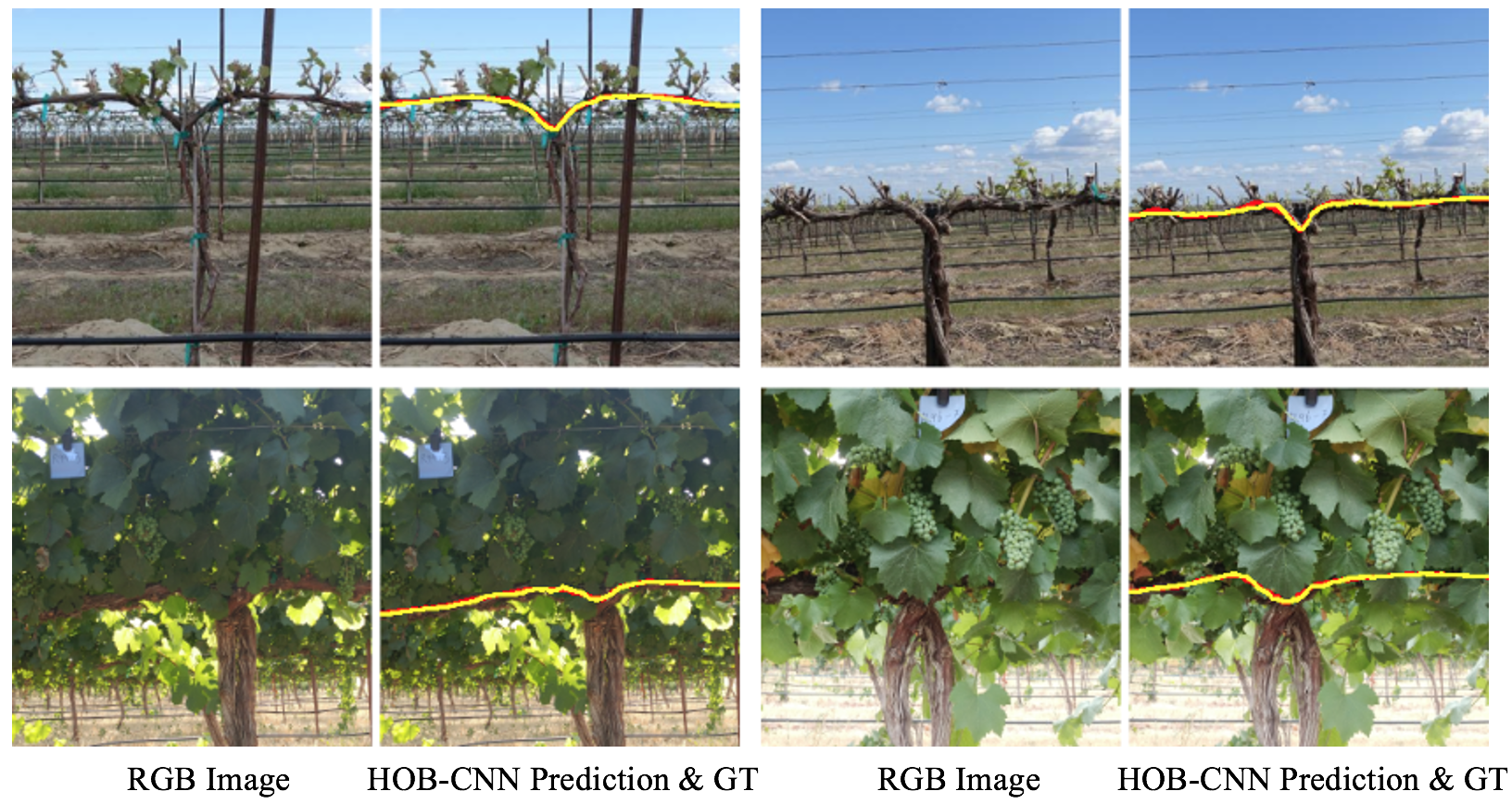}
    \caption{HOB-CNN prediction examples on Grapevine Dataset. Yellow curves are the predictions and red curves are ground truth.}
    \label{fig:grapevine}
\end{figure}

\textbf{Grapevine Dataset}: The performances of HOB-CNN at grapevine detection are shown in \cref{table:grapevine}. From early spring to summer, the RMSE increases from 2.279 to 2.661 as a result of increasing occlusions. Contrary to intuition, the correlation coefficient in summer is 0.937, which is slightly higher than that in early spring (0.925). This can potentially be explained by the larger amount and diversity of the summer dataset, which results in the improved performance on summer images. Some examples are shown in \cref{fig:grapevine}.

Trellis Tree and Grapevine Datasets are only utilized to test HOB-CNN on other 2D-structured fruit trees. Both the number and the diversity of these two datasets are limited. However, the experiments show HOB-CNN’s ability at detecting different 2D trees. In the future, more types of trees and more occlusion conditions will be considered.

\subsection{HOB-CNN Error Analysis}
\label{sec:error}

\begin{figure*}
    \centering
    \includegraphics[width=1\linewidth]{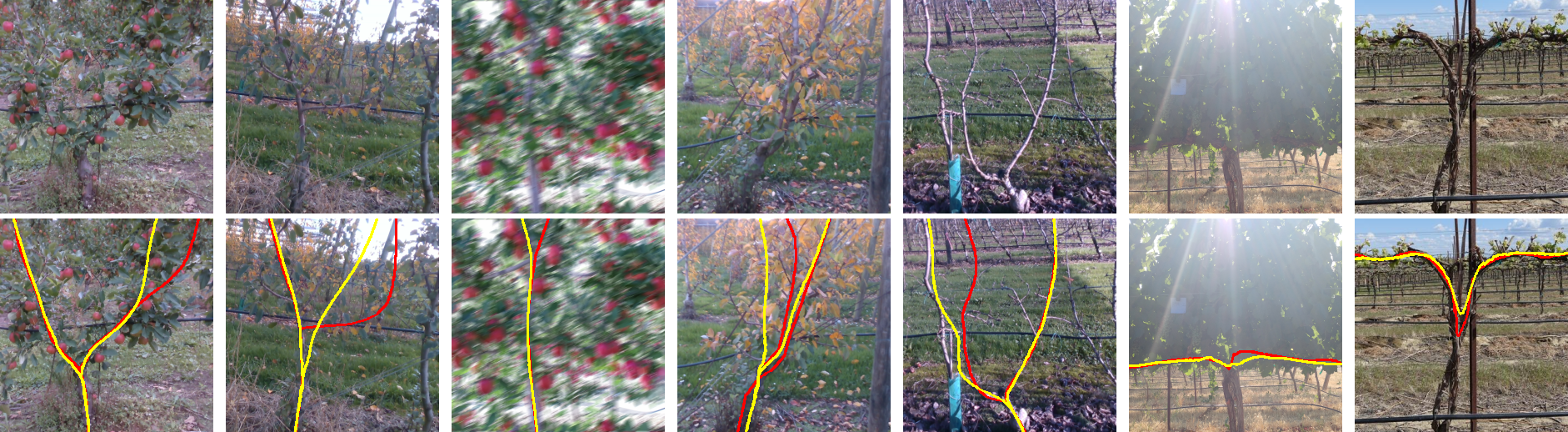}
    \caption{Examples of large error predictions. Top row: RGB images; bottom row: predictions (yellow curves) overlay with ground truths (red curves). From left to right: Extremely Occluded Tree, Thin Branch, Blurry Image, Unusual Tree Shape, Neighbouring/Background Trees, Overexposed Image and Sharp Bend Tree.}
    \label{fig:failures}
\end{figure*}

We analyze HOB-CNN predictions which have a RMSE larger than 4 pixels from all three datasets. Common features of these large error images and their corresponding image numbers are summarized in \cref{table:error} and the details are described below. The examples are shown in \cref{fig:failures}.

\begin{table}[h!]
\centering
\begin{tabular}{lccc}
\hline
\multirow{2}{*}{\textbf{Common Feature}} & \multicolumn{3}{c}{\textbf{Number of image}}                      \\ \cline{2-4} 
                                         & \textbf{Y-shaped Tree} & \textbf{Trellis Tree} & \textbf{Grapevine} \\ \hline
Extremely Occluded Tree                  & 47                     & 2                     & 43                 \\
Thin Branch                              & 7                      & 0                     & 0                  \\
Blurry Image                             & 2                      & 9                     & 0                  \\
Unusual Tree Shape                       & 16                     & 2                     & 0                  \\
Neighbouring/Background Tree             & 8                      & 0                     & 0                  \\
Overexposed Image                        & 0                      & 0                     & 3                  \\
Sharp Bend Tree                          & 27                     & 0                     & 38                 \\
Others                                   & 40                     & 2                     & 22                 \\ \hline
Total                                    & 148                    & 14                    & 106                \\
Dataset Percentage                       & 6.80\%                 & 2.40\%                & 24.40\%            \\ \hline
\end{tabular}
\caption{Common features of the large error images and their corresponding image numbers in Y-shaped Tree, Trellis Tree and Grapevine Datasets.}
\label{table:error}
\end{table}

\textbf{Extremely Occluded Tree} situations involves continues large parts of the branch being occluded. In these situations, the branch cannot be easily found due to the occlusion by the fruits and leaves in the picture. This problem might be resolved with more training data, or by matching trees against less occluded seasons. Both of these will be left to future work. 

\textbf{Thin Branch} only appears in Y-shaped trees. In these images, the thin branches often only has one or two pixels width, which makes it hard for the neural network to extract the features.

\textbf{Blurry Picture} has less detail information, resulting in inferior predictions. In practice, blurry pictures can be avoid easily by stabilizing the camera.

\textbf{Unique Tree Shape} (e.g. long trunk, close branches, sloping tree) have less images in training datasets than normal shaped trees, which contributes the bias of the vision system towards the normal shaped trees. This can be fixed by including more unique shaped trees into training dataset.

\textbf{Neighbouring/Background Tree} situations involve disturbing branches from neighbouring or background trees. The same features of the branches tricked the vision system to mistake the wrong branch.

\textbf{Overexposed Image} is similar to blurry image. Details are washed out by the highlights, and thus there are less details in the image. This can be fixed by adding polarized filter lens to the camera under strong lights.

\textbf{Sharp Bend Tree} happens in Y-shaped trees and grapevines. The prediction around the sharp bend are less accurate than around smooth branch. This can potentially be explained by the bias towards smooth branches in the training material, in which smooth branches have much more images than sharp bend branches.

\section{Conclusion}
\label{sec:conclusion}

This work proposes a novel single-step regression-based deep learning model, HOB-CNN, for predicting branch position of 2D structured fruit trees under different occlusion conditions. This work shows the potential of regression models in foliage tree branch prediction. The comparative experimental results indicate it outperforms two baselines that represent the common approaches for occluded branch hallucination with higher accuracy and better robustness to occlusions. Furthermore, HOB-CNN shows the ability to generalize across different trees. 

However, during harvest season, the occlusions on the branches can be so heavy that occasionally even humans are unable to recognize the branches on the images. Thus, it is very challenging to manually annotate the training targets, leading to the imbalanced Y-shaped Tree Dataset (more winter tree images) in this work. In future work, more data is going to be collected to make the dataset balance. Furthermore, we will train the vision system to predict the heavily occluded branches by providing an additional input of lightly occluded images of the same tree. Additionally, we will assign the thickness profile to each branch position prediction, aiming to accurately reconstruct 3D models of the tree.

\section{Acknowledgments}
\label{sec:Acknowledgments}

We thank for the support by ARC Nanocomm HUB IH150100006 and Fankhauser Apples.

\bibliography{mybibfile}

\end{document}